\documentclass[conference]{IEEEtran} 

\IEEEoverridecommandlockouts
\usepackage{cite}
\usepackage{amsmath,amssymb,amsfonts}
\usepackage{algorithmic}
\usepackage{graphicx}
\usepackage{textcomp}
\usepackage{xcolor}
\usepackage{hyperref}
\def\BibTeX{{\rm B\kern-.05em{\sc i\kern-.025em b}\kern-.08em
    T\kern-.1667em\lower.7ex\hbox{E}\kern-.125emX}}

\usepackage{amsmath,mathtools}
\usepackage{multirow,makecell}
\usepackage{booktabs}
\usepackage{placeins}

\usepackage{subcaption}

\usepackage{xcolor}

\begin{document}

\newif\ifblind

\ifblind
\else
\title{Cross-cohort TB classification using clinical data gathered in Uganda and South Africa
}
\fi

\newcommand{\orcidauthor}[2]{#1\hspace*{0.6mm}%
  \href{https://orcid.org/#2}{\includegraphics[width=0.9em]{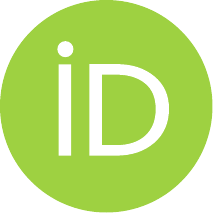}}}

\newcommand{\affilsun}{%
  \textit{Department of Electrical and Electronic Engineering},
  \textit{University of Stellenbosch}, Stellenbosch, South Africa}
\newcommand{\affilams}{%
  \textit{Department of Global and Amsterdam Institute for Global Health and Development}\\
  \textit{Amsterdam University Medical Centers, University of Amsterdam}, Amsterdam, The Netherlands
  }
\newcommand{\affilctr}{%
  \textit{South African Medical Research Council Centre for Tuberculosis
  Research (CTR)}\\
  \textit{Division of Molecular Biology and Human Genetics, Faculty of Medicine
  and Health Sciences}\\
  \textit{Stellenbosch University}, Cape Town, South Africa}
\newcommand{\affilgot}{%
  \textit{Faculty of Business and Economics, University of G\"{o}ttingen}, G\"{o}ttingen, Germany}

\newcommand{\affilmak}{%
  \textit{Department of Medical Microbiology, College of Health Sciences}\\
  \textit{Makerere University, Biomedical Research Center (MAKBRC)}, Kampala, Uganda}

\newcommand{\emailaddr}[1]{{\footnotesize\normalfont\texttt{#1}}}

\ifblind
  \author{\IEEEauthorblockN{Anonymous Author(s)}
  \IEEEauthorblockA{Affiliation withheld for double-blind review}}
\else
    \author{
    \IEEEauthorblockN{
      \orcidauthor{Joshua M. Jansen van V\"{u}ren}{0000-0002-3406-4788}\IEEEauthorrefmark{1},
      \orcidauthor{Devendra S. Parihar}{0000-0001-6404-6428}\IEEEauthorrefmark{1},
      \orcidauthor{Daphne Naidoo}{0000-0001-7711-9526}\IEEEauthorrefmark{2},
      \orcidauthor{Marisa Klopper}{0000-0002-9318-8289}\IEEEauthorrefmark{2},
    \orcidauthor{Frank Cobelens}{0000-0002-4367-1133}\IEEEauthorrefmark{3},\\
    \orcidauthor{Lutz Kolbe}{0000-0003-4852-0040}\IEEEauthorrefmark{4},
      \orcidauthor{Kimsey Zajac}{0009-0005-0385-185X}\IEEEauthorrefmark{4},
    \orcidauthor{Willy Ssengooba}{0000-0002-1643-110X}\IEEEauthorrefmark{5},
    \orcidauthor{Moses Joloba}{0000-0002-0334-9983}\IEEEauthorrefmark{5},
      \orcidauthor{Grant Theron}{0000-0002-9216-2415}\IEEEauthorrefmark{2},
      \orcidauthor{Thomas R. Niesler}{0000-0002-7341-1017}\IEEEauthorrefmark{1}}%
    \IEEEauthorblockA{\IEEEauthorrefmark{1}\affilsun}%
    \IEEEauthorblockA{\IEEEauthorrefmark{2}\affilctr}%
    \IEEEauthorblockA{\IEEEauthorrefmark{3}\affilams}%
    \IEEEauthorblockA{\IEEEauthorrefmark{4}\affilgot}%
    \IEEEauthorblockA{\IEEEauthorrefmark{5}\affilmak}%
    \IEEEauthorblockA{%
      {\footnotesize\normalfont Corresponding authors:}\hspace*{0.35em}%
      \emailaddr{jjvanvueren@sun.ac.za}, \emailaddr{trn@sun.ac.za}}%
    }
\fi

    


\maketitle

\begin{abstract}
We present a first evaluation of machine learning applied to patient clinical and demographic data gathered in two different countries for the purpose of tuberculosis (TB) screening to identify people who would benefit from expensive molecular testing.
Experiments are based on the recently-compiled CAGE-TB dataset, which includes sub-cohorts of people with presumptive TB presenting at community health care centres in South Africa and Uganda.
Three neural network architectures (logistic regression (LR), multilayer perceptrons (MLP) and convolutional neural networks (CNN)) are considered in conjunction with greedy feature selection.
For the convolutional neural network, a strategy that jointly optimises feature selection and feature ordering is proposed and shown to lead to consistent development and test set improvements.
For all three models, development set area under the receiver operating characteristic (AUROC) curve is improved by 2-7\% using feature selection.
LR after feature selection achieves an AUROC of 0.8\ [0.75,0.86] (95\% CI) and 0.84\ [0.78,0.9] when testing on the held-out Ugandan and South African data respectively.
Although outperforming LR on the development cohort, the deeper networks (MLP, CNN) show inconsistent trends on the held-out cohorts, while LR achieves performance within 1-2\% of the best achieved in terms of AUROC. 
LR narrowly misses the WHO minimum requirements by 4-9\% in sensitivity even though the network is being evaluated on a completely held-out cohort. 
The development of neural-network based classifiers for TB screening therefore appears viable. 

\end{abstract}


\section{Introduction}

Despite being treatable, TB remains a primary cause of mortality in lower and middle income countries.  
A key challenge is that many cases remain undiagnosed, allowing the disease to spread~\cite{tb_report}.
To address this, the World Health Organisation (WHO) guidelines propose an accessible TB screening tool that achieves 60\% specificity at 90\% sensitivity as part of a two step screening protocol~\cite{who_requirement}.
Such a tool should ideally be low-cost and scalable and, most importantly, provide an efficient and effective way to identify people presenting at busy primary care clinics who require costly confirmatory molecular testing.


This work describes the development of such a screening method based on easily-obtainable patient-level clinical and demographic information, and machine learning.
All the considered clinical and demographic values are either self-reported, or the result of a basic measurement, or the result of a point-of-care test.
While such systems have been proposed before, we place specific focus on the robustness of such a classifier to the domain shift of the underlying features associated with a change in the target population.
Such a cross-cohort analysis has not yet been reported.

The contributions of our work are as follows.
\begin{itemize}
    \item We present the first analysis of clinical and demographic data-based TB classification using neural networks across two different countries (South Africa and Uganda).
    \item We show that such classifiers can be robust to the associated domain shift and therefore may represent a viable approach to TB screening.
    \item We propose and show the effectiveness of a joint feature selection and ordering strategy allowing the effective application of CNN to unstructured inputs such as clinical and demographic data which do not require any laboratory-based test.
\end{itemize}


\section{Related work}

Several studies have considered the classification of clinical and demographic indicators as a way to triage for tuberculosis\cite{jaganath2024accelerating, Orjuela-Canon2025-eh, kafentzis2023predicting}.
Several of these studies consider only linear scoring strategies, or rely on a single isolated measurement~\cite{Ruperez2023,Cox2024aj,Calderwood2022,Derendinger2025CRP}.
For example, a basic linear score comprising age, sex, HIV, diabetes, cough, fever, weight loss, night sweats, and longer than 14-days of symptoms achieved an AUROC of 0.75 [0.69-0.80] in a validation cohort of $n=\ $387 adults in Uganda~\cite{Yeonsoo_2020}.

The application of machine learning methods for TB classification using datasets that include a wider set of patient-level clinical and demographic information is an area of growing interest.
A random forest model applied to classify the development of TB within a six month time-frame for people living with HIV based on 48 laboratory and demographic indicators was shown to achieve an AUROC of 0.83~\cite{Bartl2025-jo}. 
Similarly, various sets of clinical data were considered as inputs for machine learning TB classifiers in 
\cite{Orjuela-Canon2025-eh} and \cite{botha2018}, achieving AUROCs of 0.7 and 0.8 respectively.
Generally, higher performance is reported when laboratory based tests are included as inputs to the machine learning models.
For example, utilising eighteen features derived from complete blood counts, an XGBoost classifier is shown to achieve an AUROC of 0.91 when trained to classify people with TB~\cite{Ghermi2024-zv}. 
However, reliance on laboratory tests reduces the accessibility of such methods.

In related work developing a neural network TB classifier based on cough audio, an AUROC of 0.74 [0.70,0.78] was achieved~\cite{jaganath2024accelerating}.
This was, however, outperformed by a model trained only on clinical data (AUROC 0.81 [0.78,0.85]).
Furthermore, through external validation of the cough audio based models to cough data from Peru, which is not one of the countries in which data from the original study was collected, AUROC fell to between 0.48 and 0.62~\cite{Zimmer2026}.
This highlights the challenge of developing a tool which is robust to application in unseen settings as well as the necessity of external validation for TB screening.

Studies such as those mentioned above have considered the overall TB classification performance that can be achieved by machine learning using various clinical and demographic data in various settings.
However, no study has systematically examined how these classifiers generalise when applied to a held-out cohort from another country, where demographic and clinical features may be differently distributed due to geographic differences in risk for contracting TB and potential differences in health systems and clinical practices.
We address this gap by conducting a cross-cohort TB classification analysis for two countries, Uganda and South Africa.
The results we obtain can be compared with our previous work, in which only South African data was considered~\cite{Parihar2025}.

\section{Data}
\label{sec:data}

We utilise data from the Cough Audio triaGE for TB (CAGE-TB) dataset, which was recently compiled in South Africa and Uganda.
Each country sub-cohort includes clinical and demographic data collected from people who have self-presented to a community clinic and are symptomatic for TB.
We list the 28 clinical and demographic features that were successfully collected for every such person in Table~\ref{tab:data_cage}.
The enrolment criteria for the study included that a person should have been experiencing a cough lasting at least two weeks.
No participants experiencing haemoptysis (coughing up blood) were enrolled.

\begin{table}[h!]
\centering
\caption{Clinical and demographic data gathered for each participant in the data used for experimentation.}
\label{tab:data_cage}
\setlength{\tabcolsep}{4pt} 
\begin{tabular}{l l}
    \toprule
    Feature & Type \\
    \midrule
%
    Age                           & Quantitative (years) \\ 
    Sex                           & Binary (\texttt{female, male}) \\
    Dyspnea                       & Binary (\texttt{yes, no}) \\
    Chest pain                    & Binary (\texttt{yes, no}) \\
    Previous TB                   & Binary (\texttt{yes, no}) \\
    Night sweats                  & Binary (\texttt{yes, no}) \\
    Chronic conditions            & Binary (\texttt{yes, no}) \\
    Fever within last two weeks   & Binary (\texttt{yes, no}) \\
    Coughing up sputum            & Binary (\texttt{yes, no}) \\
    Recent weight loss ($>$1.5kg) & Binary (\texttt{yes, no}) \\
    Anaemic conjunctivae          & Binary (\texttt{yes, no}) \\
    HIV status                    & Binary (\texttt{yes, no})  \\

    Height                        & Quantitative (cm)  \\
    Weight                        & Quantitative (kg)   \\
    BMI                           & Quantitative (kg/m\textsuperscript{2}) \\
    CRP concentration             & Quantitative (mg/L)\\
    Mid-upper arm circumference& Quantitative (cm) \\
    Haemoglobin concentration     & Quantitative (g/dl)\\
    Cough duration                & Quantitative (days)  \\  
    Number of symptoms            & Quantitative (\texttt{1,2,3,4})  \\
    Years since previous TB       & Quantitative (years) \\
    TBscoreII~\cite{Rudolf2013_tbscore_II} & Quantitative (\texttt{1,2,$\dots$,8}) \\

    \multirow{2}{*}{ART status}   & Multinomial (\texttt{no - HIV neg,} \\
                                  & \texttt{no - HIV pos, yes - HIV pos}) \\
    Respiratory rate              & Quantitative (breaths/min)  \\
    Heart rate                    & Quantitative (BPM) \\ 
    Axillary/ear temperature      & Quantitative (Celsius) \\
 \multirow{2}{*}{Smoking}         & Multinomial (\texttt{current, never,} \\
                                  & \texttt{stopped <1,<12,>12 months}) \\
    Fagerstrom score              & Quantitative (\texttt{-6,0,1,...,10})  \\ 

\bottomrule
\end{tabular}
\end{table}





Of the CAGE-TB South African and Ugandan sub-cohorts, we consider only those participants for whom all 28 clinical and demographic measurements listed in Table~\ref{tab:data_cage} are available.
Records with missing values are therefore disregarded in the work presented here.
Under this constraint, the South African sub-cohort includes 347 participants, and the Ugandan sub-cohort 374 participants.
The number of TB positive and TB negative cases per cross-validation fold in each sub-cohort is presented in Table~\ref{tab:cross_cohort_stats}. 
The TB status of each individual was determined using a composite reference standard comprising either a MGIT 960 liquid culture or an Xpert MTB/RIF Ultra positive result.

\begin{table}[h!]
    \caption{TB positive and negative participants per cross validation fold for the South African (SA) and the Ugandan (UG) sub-cohorts used for experimentation.}
    \label{tab:cross_cohort_stats}
    \setlength{\tabcolsep}{4pt}
    \centering
    \begin{tabular}{c c c c c c c c c c c c c}
        \toprule
        Country & Fold & 1 & 2 & 3 & 4 & 5 & 6 & 7 & 8 & 9 & 10 & \textbf{Total} \\
        \midrule
        \multirow{2}{*}{SA} & \# Pos & 5& 6& 8& 6& 7& 8& 6& 7& 9& 7 & 69 \\
                            &  \# Neg  & 29& 27& 33& 29& 25& 27& 31& 28& 24& 25 & 278 \\
        \cmidrule{2-13}
        \multirow{2}{*}{UG} & \# Pos  & 10& 8& 8& 10& 9& 9& 8& 9& 10& 13  & 94 \\
                            &  \#  Neg  & 31& 29& 22& 32& 31& 30& 24& 25& 25 & 31 & 280 \\
        \cmidrule{2-13}
        \multirow{2}{*}{\textbf{Total}} & \# Pos    & 15& 14& 16& 16& 16& 17& 14& 16& 19& 20 & 163 \\
                             & \# Neg      & 60& 56& 55& 61& 56& 57& 55& 53& 49& 56 & 558 \\
        \bottomrule
    \end{tabular}
\end{table}

\section{Experimental setup}
\label{sec:exp_setup}

In the following we describe the classifier architectures used for experimentation, the optimisation strategy for the associated hyperparameters, the configuration of the cross-cohort experiments, the way in which the clinical and demographic features are encoded for use as classifier inputs, and the method of feature set optimisation.
The results of the experiments themselves are then presented in Section~\ref{sec:results}.

\subsection{Neural network classifiers}

We employ three different neural network architectures: logistic regression, multi-layer perceptron, and a one-dimensional convolutional neural network.

\vspace*{1mm}
\subsubsection{Logistic regression} 
\label{sec:logistic_regression}

The formulation of LR,\ is presented in Equation~\ref{eq:logistic_regression}, where $\boldsymbol{x}$ is the input encoded feature vector while $\boldsymbol{w}$ and $b$ are the learned weight vector and scalar bias respectively. 

\begin{equation} \label{eq:logistic_regression}
    \hat{y} = \frac{1}{1+e^{-(\boldsymbol{w}^T \boldsymbol{x} + b)}}
\end{equation}

The model is trained to let  $\hat{y}$ approximate the posterior probability of TB positivity. 
LR is a linear model with comparatively few parameters.
It therefore has the advantage of being less likely to overfit, but is limited by its linear decision surface.

\vspace*{1mm}
\subsubsection{Multi layer perceptron} 
\label{sec:mlp}

The MLP architecture comprises several fully connected layers, each with an activation function.
The first layer receives the feature vector as input, and the activations of this layer are passed to the subsequent layers.
The activations of all except the final layer are often referred to as \textit{hidden} representations.
The final layer produces a binary output, again approximating the posterior probability for TB.
In comparison with LR, MLPs are able to model non-linear relationships in the training data.
However, they are also more prone to overfitting, and therefore
only two and three layer networks are considered here. 
The dimensionality of the hidden representations, as well as the type of activation function in the hidden layers are treated as hyperparameters and optimised using a grid-search during cross-validation.
The values found in this way are presented in Table~\ref{tab:hyperparameters}.

\vspace*{1mm}
\subsubsection{Convolutional neural network} 
\label{sec:1d_cnn}

Previous work has indicated that a CNN can be applied to clinical and demographic data for tuberculosis classification~\cite{Parihar2025}.
Here we apply the same model architecture, as presented in Table~\ref{tab:1d_cnn_structure}.
As for the LR and MLP classifiers, the network is trained to approximate the posterior probability for TB.

\begin{table}[h!]
\centering
\caption{One-dimensional CNN structure reproduced from~\cite{Parihar2025}. The output dimensions of all but the last layer depend on the input dimension.}
\label{tab:1d_cnn_structure}
\setlength{\tabcolsep}{4pt}
\begin{tabular}{ c c c c c c}
    \toprule
    \# & Layer type & Kernel dim & Num ($C$) & Stride ($s$) & Activation \\
    \midrule
    1 & Conv &  $(3 \times 1 )$  & 20 & 1    & ReLU  \\
    2 & Max pool &  $(2 \times 1 )$     & - & 1     & - \\
    3 & Conv &  $(3 \times 1 )$  & 50 & 1    & ReLU \\
    4 & Max pool &  $(2 \times 1 )$     & - & 1     & - \\
    \midrule
    5 & Dense & - & - & - & Sigmoid \\
    \bottomrule
\end{tabular}
\end{table}

Convolutional neural networks are typically applied to images, because they have been shown to very effectively leverage the identification of local features to achieve strong classification performance.
Images encode the specific geometric structure of the objects they contain in the spatial ordering of the pixels.
However, no such spatial structure or defined ordering exists for the kind of demographic and clinical data we consider here.
Therefore, for a CNN to be effective, the ordering of the feature vector must be optimised.
In previous work this was achieved through the brute-force repeated retraining of the model using different random permutations of the ordering.
In Section~\ref{sec:supp_sfs} we will propose a new and more effective strategy for discovering the best way to arrange and order the feature vector.

The closest related research considering the application of CNNs to tabular inputs in which a natural feature ordering is absent is found in the area of genomics and transcriptomics.
In this case, genetic tabular information is arranged into a two-dimensional grid for application to image-based CNNs.
This can be accomplished either by preserving inter-feature distances~\cite{Zhu2021,Bazgir2020,Sharma2019} or by learning the assignment end-to-end~\cite{MATSUDA2024111293}.
Alternatively, previous work has investigated inserting a learned dense projection ahead of a one-dimensional CNN so that the convolution operates on a soft-reordering of the input features rather than on the raw features~\cite{QIAN2023110414}.
In contrast, in the work we present here we retain a convolution over the raw feature vector, and perform an explicit joint search over feature selection and positional ordering in one dimension to directly optimise the development-set AUROC.

\subsection{Hyperparameter search} 
\label{sec:hyperpara_optim}


We search the set of hyperparameters presented in Table~\ref{tab:hyperparameters} in a grid-fashion, selecting those which optimise the average development set AUROC.
The hyperparameters for each of the three architectures are optimised separately for the Ugandan and for the South African sub-cohort.
In addition, the hyperparameter search is repeated for both the full set of 28 input features as well as for the reduced set of input features identified during sequential feature selection, as described in Section~\ref{sec:sfs}.
The Adam optimiser is utilised in all experiments, with a weight decay of 0.01~\cite{kingma2014adam}.
Weight decay was considered a hyperparameter in preliminary experiments, and a value of 0.01 was found to be optimal for all three models.
All networks are trained using PyTorch~\cite{pytorch}.

\begin{table}[h!]
\centering
\caption{Hyperparameters and the ranges considered during optimisation. Optimal hyperparameters for LR and the MLP are presented for the subset of features identified during SFS. Optimal parameters for the CNN are presented after the joint optimisation of the order and chosen features of the input vector. SA: Optimal hyperparameters for South Africa, UG: Optimal hyperparameters for Uganda.
}
\vspace{2mm}
\label{tab:hyperparameters}
\setlength{\tabcolsep}{2pt}
\begin{tabular}{l l c c c}
    \toprule
    Classifier & Hyperparameter & Range & SA & UG \\
    \midrule
             LR      & Learning rate       & 0.01, 0.001                                  & 0.01          & 0.001 \\
             LR      & Batch size          & 8, 16, 32                                    & 8             & 32     \\
             LR      & Epochs              & 1,$\dots$,128                                & 2             & 66     \\
             \midrule
             \multirow{2}{*}{MLP}    & \multirow{2}{*}{Activation} & ReLU, Tanh             & \multirow{2}{*}{Sigmoid} & \multirow{2}{*}{Sigmoid} \\
                                        &                          & Sigmoid                &        \\
             \multirow{3}{*}{MLP}       & Hidden layer & [16], [64]                         &        \\
                 & dimensionality                      & [16,16], [16,64]                   &  [64] & [16]      \\ 
                 & $\left[\text{Layer 1},\dots, \text{Layer }n \right]$ & [64,16], [64,64]  &        \\
                                                                                                     
             MLP    & Learning rate         & 0.01, 0.001, 0.0001                           & 0.01 & 0.01   \\
             MLP    & Batch size            & 8, 16, 32, 64 , 128                           & 32 & 32       \\
             MLP    & Dropout rate          & 0.1, 0.3, 0.5, 0.7                            & 0.3 & 0.3     \\
             MLP      & Epochs              & 1,$\dots$,128                               & 3 & 89        \\
             \midrule
             CNN     & Learning rate       & 0.01, 0.001                                  & 0.001          & 0.01\\
             CNN     & Batch size          & 8, 16, 32                                    & 16             & 16  \\
             CNN     & Epochs              & 1,$\dots$,128                                & 77             & 18  \\
    \bottomrule
\end{tabular}
\end{table}

\subsection{Cross-cohort analysis} \label{subsec:cross_cohort_analysis}
\begin{figure*}[ht!]
    \centering
    \includegraphics[width=\linewidth]{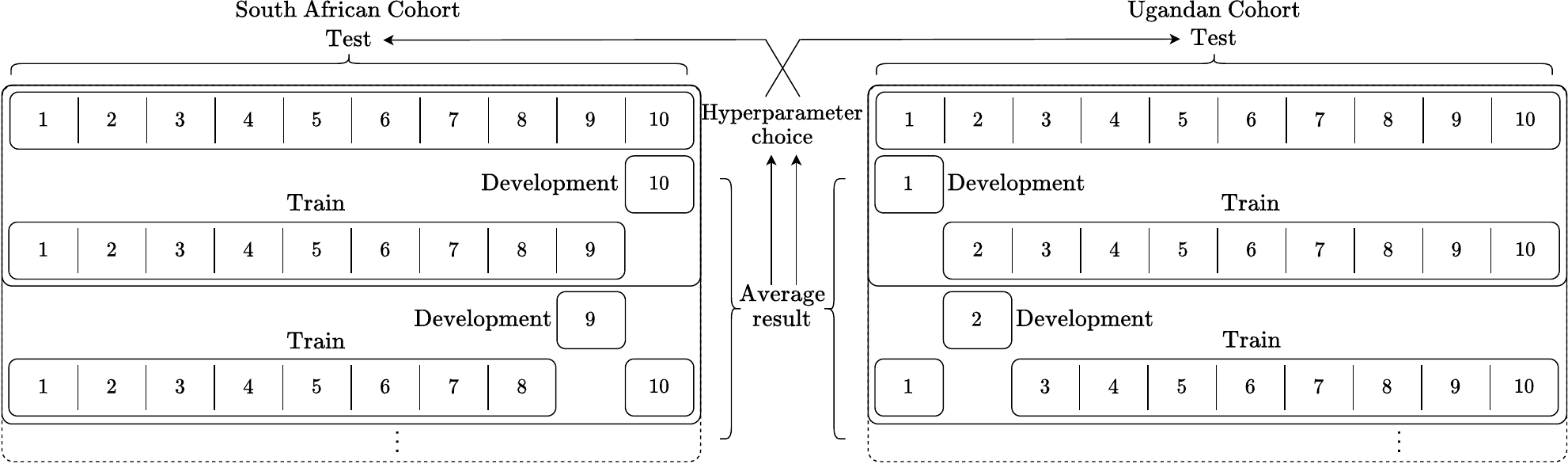}
    \caption{Illustration of ten-fold cross-cohort cross validation. In the cross validation loop, each of the ten folds is held out in turn as a development  set for hyperparameter optimisation, while the entire held-out cohort is used as a test set. }
    \label{fig:cross_validation}
\end{figure*}

Experiments are conducted by considering only one sub-cohort for model training and hyperparameter optimisation, after which the optimised models are directly applied to the other sub-cohort, which serves as a held-out test set in a different domain (country).
Subsequently, the alternate experiment in which the sub-cohorts are switched is performed. 
This procedure is illustrated in Figure~\ref{fig:cross_validation}.

Considering each cohort as a fully held-out test set serves as a means of verifying that the classifiers do not overfit the training population.
For each test-set a confidence interval is calculated using DeLong's method~\cite{delong_ci}.
Finally, because these classifiers are intended to be used for screening, an operating point must be determined.
To accomplish this, we select the classification threshold which achieves a 90\% sensitivity for each of the development sets (after hyperparameter optimisation)~\cite{who_requirement}.
This value is then averaged and applied directly to the held-out sub-cohort, from which a specificity and sensitivity is reported.

\subsection{Feature encoding} 
\label{sec:feature_encoding}

In this section we describe how the features were encoded for presentation as input to the neural network classifiers.
All quantitative values are presented as inputs without modification.
Binary features were encoded using a value of zero to indicate false, and a value of one to indicate true.
Finally, a one-hot representation was utilised for multinomial values.
Per-feature mean and variance normalisation, based on the training set, is applied during training and testing.
 
Previous work has shown that grouping chronic conditions into a single binary indicator afforded small but consistent improvements over a separate encoding~\cite{Parihar2025}.
Therefore we adopt this strategy here.
A single binary indicator is employed to encode whether the participant has suffered from any chronic condition.
Specifically this condition must have: lasted for one year or longer, require ongoing medical attention, or limit activities of daily living.

We also explore the impact of smoking habits (Table~\ref{tab:data_cage}).
This is achieved via a multinomial encoding that indicates whether a participant has: never smoked, currently smokes, has quit smoking less than one month ago, or between one and 12 months ago, or more than one year ago.

\subsection{Sequential feature selection}
\label{sec:sfs}

For both sub-cohorts, the same 28 clinical and demographic features are available (Table~\ref{tab:data_cage}).
We utilise sequential feature selection (SFS) to identify those features among these 28 which are the most useful for TB classification.
SFS is a greedy algorithm which begins by identifying the single feature which produces the best model performance, here assessed using AUROC.
It then iteratively adds additional features one by one from the remaining pool.
Each additional feature is selected based on whether its addition affords the best AUROC on average across the ten development sets during cross-validation.

We first apply SFS in this way to the South African sub-cohort, resulting in a set of features identified as optimal for this data. 
After re-introduction of the development data, this set of features is used to train a final model using the South African data.
This model is then applied to and evaluated against the Ugandan sub-cohort.
Subsequently, the corresponding alternative experiment is performed, where SFS and model training is performed on the Ugandan sub-cohort, and this is evaluated on the South African data.

\subsection{CNN sequential feature selection and ordering}
\label{sec:supp_sfs}

We propose a new strategy for simultaneously optimising feature selection and feature ordering for the CNN architecture.
Previous work directly applied features selected by a logistic regression model during SFS and then separately optimised the ordering through repeated random shuffling the input vector~\cite{Parihar2025}.
The strategy we propose here begins with a vector of zeros, whose dimensionality is the full number of input features.
A single input feature is then inserted into the first position in the vector, and the model is trained and re-evaluated.
This occurs in the inner cross-validation loop, after which an average development set performance is calculated.
The same input feature is then moved to the second position in the input vector, and the procedure is repeated.
This continues for all positions and for all input features, allowing the best feature and its associated position in the vector to be determined and fixed, based on the average development set AUROC.
This procedure is iterated, each time selecting the next feature from the pool of yet unselected features and its optimal position, until all positions in the vector have been filled.
Unlike previous applications which have applied either a soft-reordering~\cite{QIAN2023110414}, or learn a 2D representation of the entire feature set\cite{Zhu2021,Bazgir2020,Sharma2019,MATSUDA2024111293}, our proposed greedy-approach identifies a feature ordering and simultaneously identifies a subset of features which are beneficial for classification.

\section{Results}
\label{sec:results}

\begin{table}[t!]
    \centering
    \caption{Classifier average development ($\pm$ stddev) and test (95\% CI DeLong~\cite{delong_ci}) performance. Spec @ Sens: Test set specificity and sensitivity from threshold determined to achieve 90\% sensitivity on average over the development set. Development set values are presented as mean and standard deviation over cross validation folds.}
    \begin{tabular}{l l l l l l l }
        \toprule
        \multirowcell{3}{Model} & \multirowcell{3}{South Africa \\ Dev AUROC} & \multicolumn{2}{c}{Uganda Test} \\
        \cmidrule{3-4}
        &  & AUROC & Spec @ Sens \\
        \midrule
        LR          & 0.84$\pm$0.08 & 0.77 [0.72,0.82] & 0.23 @ 0.97 \\        
        LR (SFS)    & 0.87$\pm$0.05 & 0.80 [0.75,0.86] & 0.62 @ 0.81 \\  
        CNN         & 0.84$\pm$0.06 & 0.72 [0.66,0.79] & 0.55 @ 0.78 \\       
        CNN (SFS)   & \textbf{0.91$\pm$0.03} & 0.76 [0.71,0.82] & 0.61 @ 0.80 \\ 
        MLP         & 0.88$\pm$0.04 & \textbf{0.82 [0.77,0.87]} & 0.36 @ 0.94 \\       
        MLP (SFS)   & \textbf{0.91$\pm$0.06} & 0.79 [0.74,0.85] & 0.71 @ 0.73 \\ 
        \midrule
        & \multirowcell{3}{Uganda \\ Dev AUROC} &  \multicolumn{2}{c}{South Africa Test} \\ 
        \cmidrule{3-4}
        &  & AUROC & Spec @ Sens \\
        \midrule
        LR          &  0.84$\pm$0.1 & 0.85 [0.79,0.91] & 0.85 @ 0.71 \\
        LR (SFS)    &  0.86$\pm$0.08 & 0.84 [0.78,0.9] & 0.66 @ 0.86 \\
        CNN         &  0.83$\pm$0.09 & 0.79 [0.72,0.86] & 0.74 @ 0.77 \\
        CNN (SFS)   &  0.88$\pm$0.08 & 0.83 [0.77,0.89] & 0.95 @ 0.39 \\
        MLP         &  0.89$\pm$0.06 & \textbf{0.86 [0.80,0.91]} & 0.9 @ 0.68 \\
        MLP (SFS)   &  \textbf{0.91$\pm$0.07} & 0.84 [0.79,0.90] & 0.8 @ 0.72 \\        
        \bottomrule
    \end{tabular}

    \label{tab:all_results}
\end{table}


Development and test set classifier performance in terms of AUROC are presented for both countries in Table~\ref{tab:all_results}.
The bottom half of the table presents the performance of classifiers that were trained on the Ugandan sub-cohort and tested on the South African sub-cohort without any fine-tuning.
Although trained and developed on data solely from Uganda, the performance of the classifiers presented here does not degrade on the held-out South African data compared to previous work, which reported a mean test set AUROC of 0.83 with a 95\% confidence interval of [0.79, 0.87] using fully-nested cross-validation~\cite{Parihar2025}.

On average the South African test AUROC appears slightly higher than for Uganda, but for the MLP after feature selection, which represents the model which achieved the best development set performance for both cohorts, the difference is only 5\% absolute and the confidence intervals overlap, although the populations in each are distinct.
This is important because, when datasets are limited such as in the scenario we consider, neural classifiers often do not generalise well to data from a different domain.

\begin{figure}[t!]
    \centering
    \includegraphics[width=\linewidth]{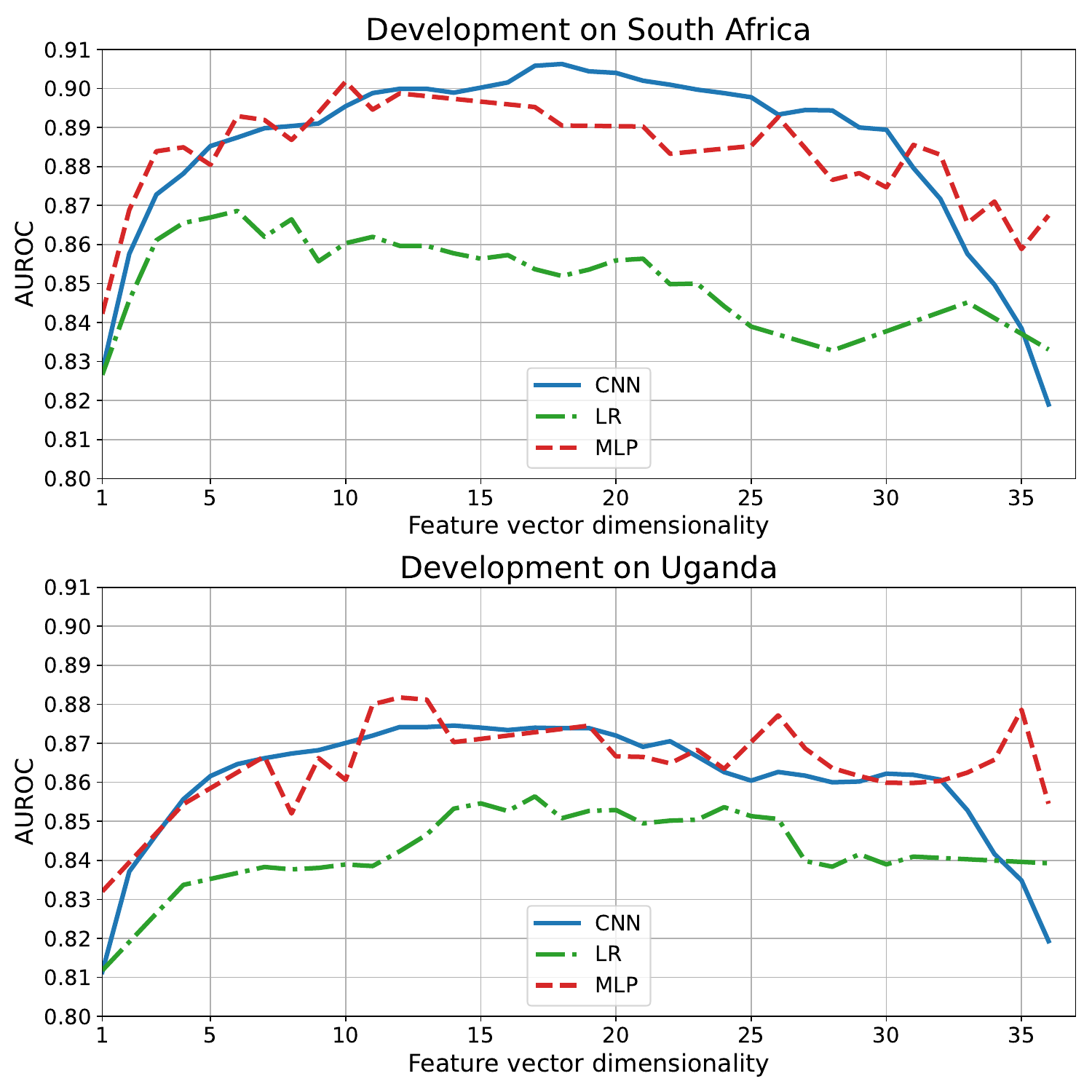}
    \caption{Average development set AUROC for each of the three considered architectures (LR: Logistic regression, MLP: Multi-layer perceptron, CNN: Convolutional neural network) during sequential feature selection (SFS) for the South African cohort (top) and Ugandan cohort (bottom).
    Horizontal axis denotes the dimensionality of the feature vector.
    }
    \label{fig:sfs_results_cross_cohort}
\end{figure}

Figure~\ref{fig:sfs_results_cross_cohort} presents the development set AUROCs for the cross-cohort SFS results for each of the three models.
The application of feature selection improves development set performance for both countries.
The joint greedy optimisation of the selected features and their position (Section~\ref{sec:supp_sfs}) for the CNN model appears to be effective, improving development set performance by 7\% and 5\% on the South African and Ugandan sub-cohorts respectively.
In addition, when considering the held-out test data, consistent improvements (of approximately 4\%) are seen for the CNN, while for LR and MLP the application of SFS achieves mixed results.

The best development set performance is consistently achieved by the MLP after feature selection, achieving an absolute AUROC of 0.91 for both cohorts, outperforming LR by between 4-5\% absolute in development set AUROC.
Test set performance, however, is similar to that achieved by the much simpler LR model.
Specifically, LR (SFS) outperforms the MLP by 1\% absolute in AUROC when testing on the Ugandan sub-cohort while equal performance is achieved for the South African sub-cohort.
Therefore, for these cohorts, the application of the simple logistic regression model appears favourable.
If the dataset could be enlarged, the deeper models may show further potential.

\begin{figure}[ht!]
    \centering
    \includegraphics[width=\linewidth]{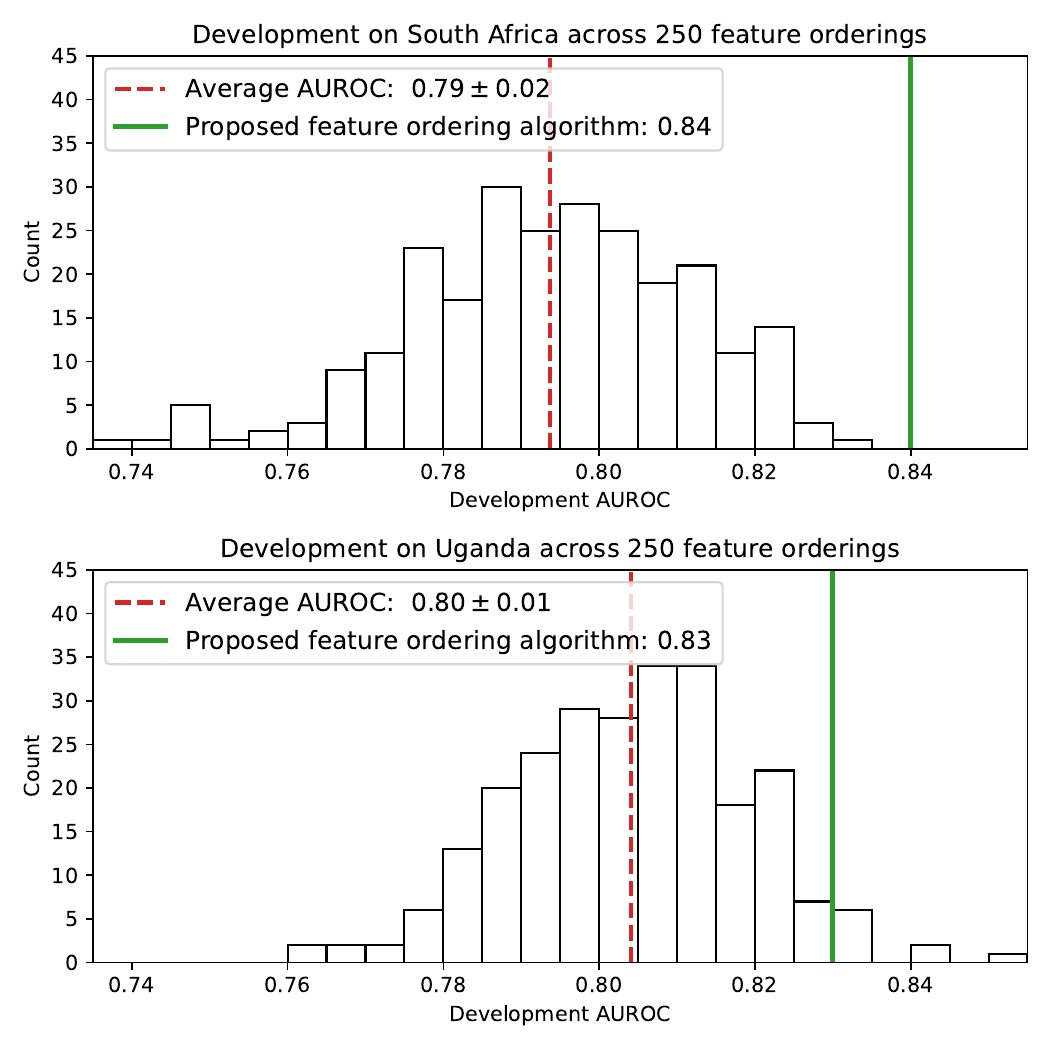}
    \caption{Average development set AUROC over 250 CNN retraining runs, each using a different random ordering of the full input vector. Bins widths are 0.5\%.
    }
    \label{fig:dev_performance_cnn_approach}
\end{figure}

In Figure~\ref{fig:dev_performance_cnn_approach} we show the average development set AUROC achieved after 250 random permutations of the feature vector ordering for the CNN.
The distribution shows that the ordering has a substantial impact on CNN classification performance, with a total spread of approximately 9\% absolute in AUROC.
For both countries, the proposed joint feature selection and ordering optimisation algorithm identifies an ordering which far surpasses the average performance of the random permutations.
This pattern is even stronger when the feature ordering is selected based on the results of SFS with LR.
After selecting the ordering which achieved the best average development performance of the 250 variants, the random search yields an AUROC of 0.87 and 0.86 for the South African and Ugandan development sets respectively, trailing our proposed search strategy by 4\% and 2\% absolute.

The optimal hyperparameters when utilising features selected by SFS were presented in Table~\ref{tab:hyperparameters}.
For both cohorts, the optimal MLP hyperparameters indicate that a small two layer model is preferred, despite deeper networks being searched.
Again, it therefore appears that, given the small size of the corpus, more regularised models such as logistic regression remain a better choice.

\subsection{Feature ordering}

\begin{figure*}[ht!]
    \centering
    \begin{subfigure}[b]{0.45\textwidth}
        \centering
        \includegraphics[width=\linewidth]{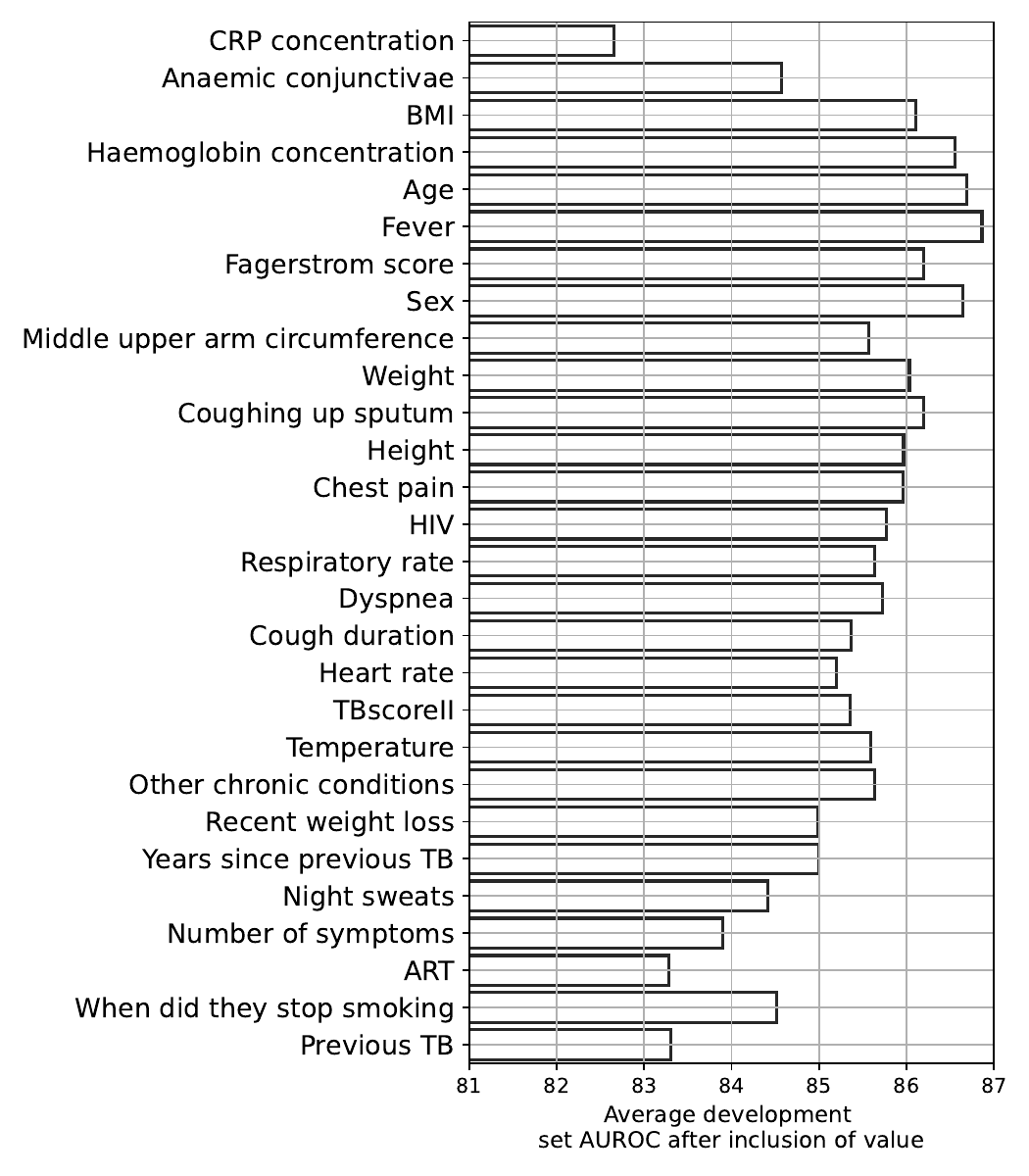}
        \caption{South African development set.}
    \end{subfigure}
    \begin{subfigure}[b]{0.45\textwidth}
        \centering
        \includegraphics[width=\linewidth]{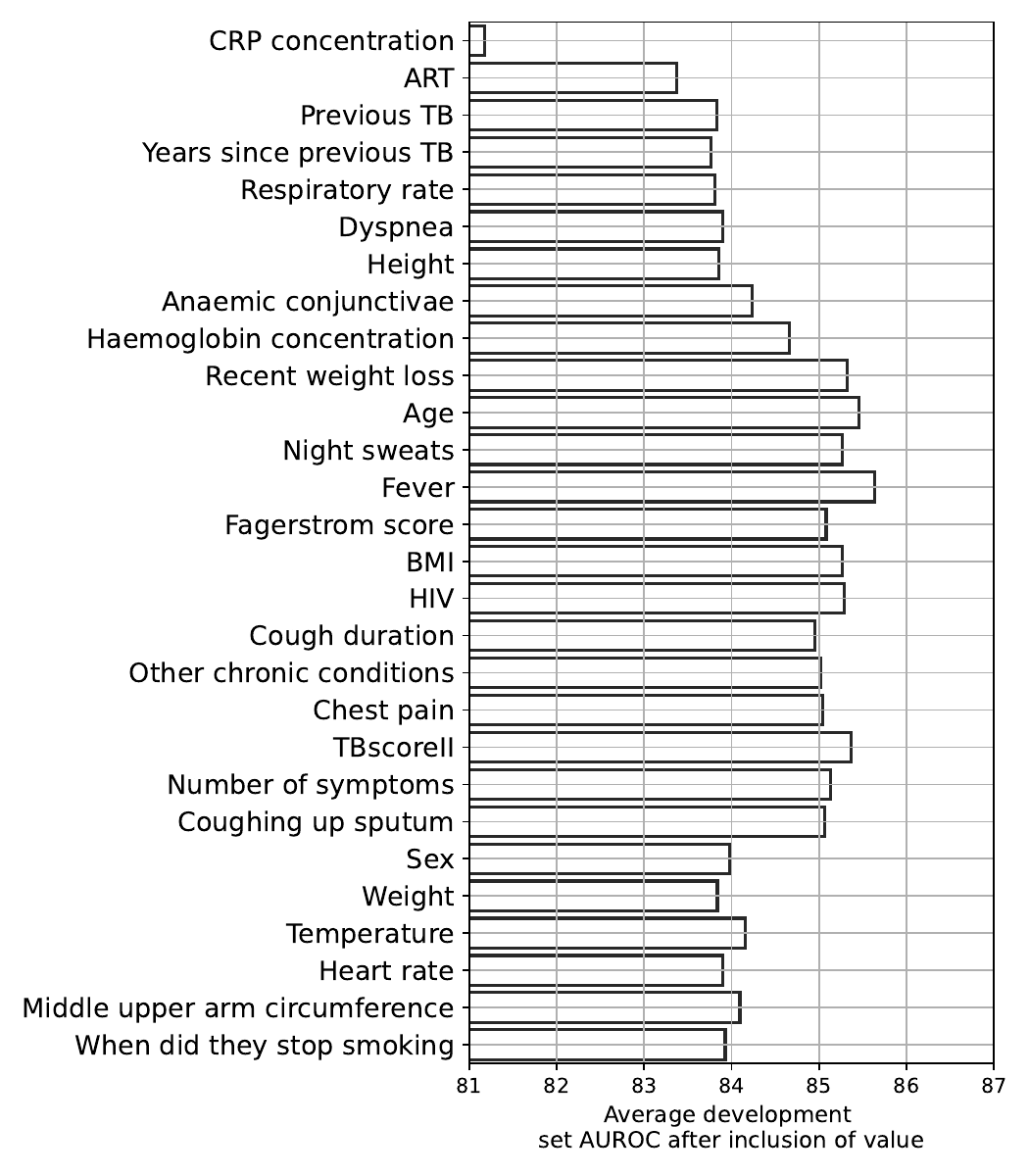}
        \caption{Ugandan development set.}
    \end{subfigure}
    \caption{Average development set AUROC after including each variable during sequential feature selection for either the South Africa (a)  or Ugandan (b) cohorts.} 
    \label{fig:cros_country_sfs_features}
\end{figure*}

Given the variability across development and test set results seen for the deeper models (CNN, MLP) we elect to interrogate the features selected by SFS when used with logistic regression.
Figure~\ref{fig:cros_country_sfs_features} shows that for the South African cohort, optimal performance is achieved when including just six of the possible 28 features: CRP, anaemic conjunctivae, BMI, haemoglobin concentration, age and fever.
For the Ugandan cohort, on the other hand, thirteen features are selected: CRP, ART status, previous TB, years since previous TB disease, respiration rate, dyspnea, height, anaemic conjunctivae, haemoglobin concentration, recent weight loss, age, night sweats and fever.
Among these, CRP, anaemic conjunctivae, age, haemoglobin concentration, and fever occur in both sets.
Although CRP is always chosen first, and requires a point-of-care tests, an additional $\approx\ $4\% improvement in AUROC is achieved by leveraging the additional features, all of which are easily obtainable through questionnaire, or simple measurement.
For both cohorts, neither knowledge of chronic conditions nor of smoking habits had any positive impact on TB classification.

\subsection{WHO target product profile for TB screening}

In 2025, the World Health Organization (WHO) released an updated target product profile for TB screening tests which recommended a minimum specificity of 60\% at a sensitivity of 90\%.
This tool could then form the first of a two-stage screening process, where individuals positively identified in the first are recommended for further testing in the second, for example using microbiological tests or chest X-rays~\cite{who_requirement}.
This strategy aims to minimise the number of people without TB who are sent for expensive confirmatory testing.


\begin{figure}[h!]
    \centering
        \includegraphics[width=0.5\textwidth]{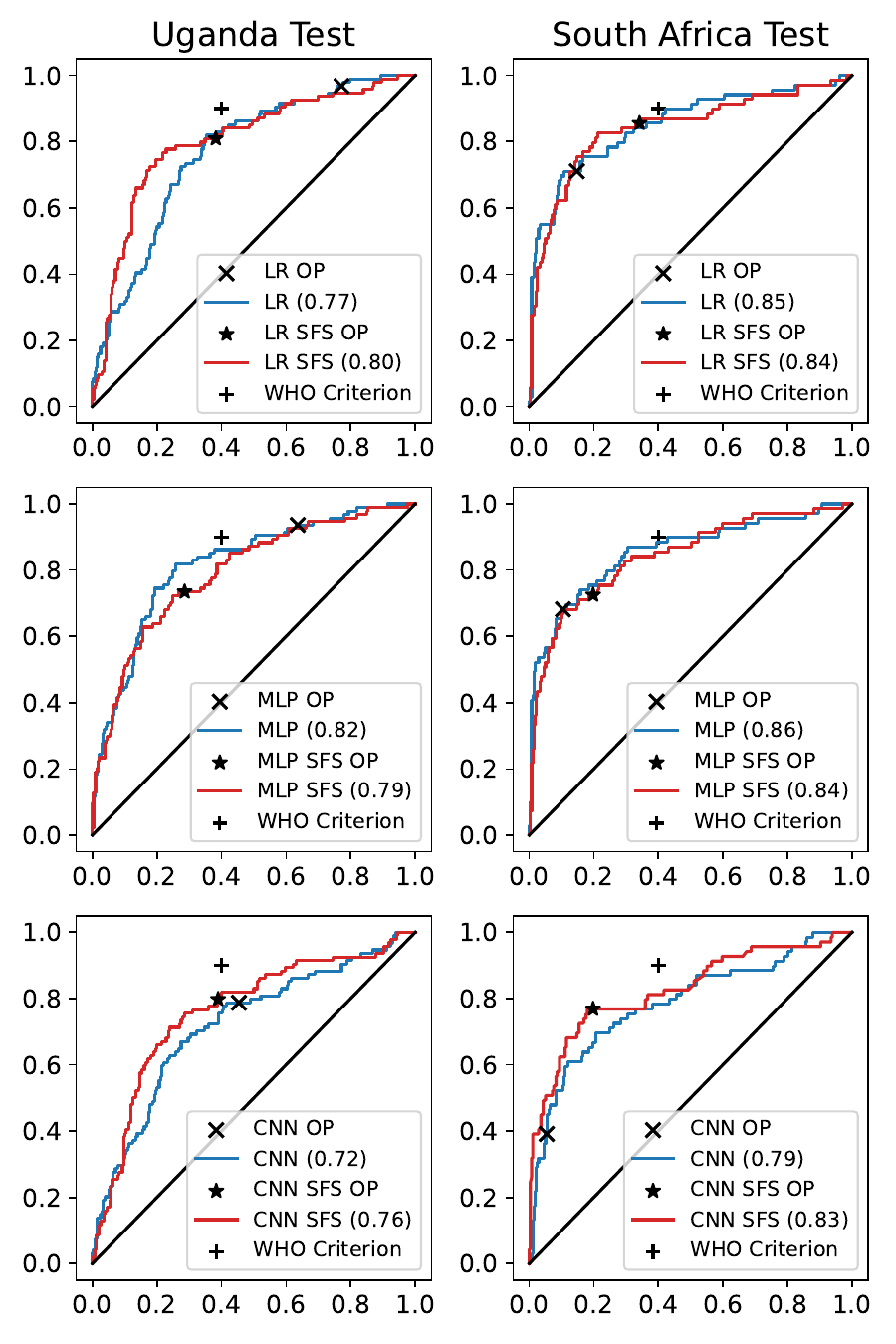}
    \caption{Neural network cross-cohort performance.
    Models are trained and thresholds determined on one cohort and tested on the held-out cohort.
    Stars ($\star$) and crosses ($\boldsymbol{\times}$) represent classifier operating points with and without SFS respectively after optimising the threshold on the development set to obtain 90\% sensitivity.
    Plus ($+$) represents the minimum WHO requirements for a screening test.
    OP: Operating point.
    }
    \label{fig:classifier_roc_curves}
\end{figure}

Figure~\ref{fig:classifier_roc_curves} presents the ROC curves for each of the classifiers, both utilising all 28 features  and those features identified by SFS.
If we consider the operating points determined on the development sets which aim to satisfy the 90\% sensitivity requirements for the neural network classifiers (Section~\ref{subsec:cross_cohort_analysis}), marked as $\star$ (SFS) and $\times$ (before SFS) in the figure, the thresholds determined on the Ugandan development cohort appear to produce performance with specificity well above the minimum requirement at the expense of sensitivity on the South African test set.
The opposite appears to be true for the classifiers developed on the South African data and evaluated on the Ugandan sub-cohort.
This highlights the challenge of applying a developed neural network classifier to a held-out setting, and may substantiate the need for network calibration and potentially additional data collection from the target area to determine a cohort specific threshold.
This inconsistency is seen for all classifiers except for LR after SFS.

The model which achieves the most consistent performance across both cohorts is LR after applying SFS.
This model narrowly misses the WHO minimum specifications for a screening tool by 4\% in terms of sensitivity (86\% sensitivity at specificity 66\%) when tested on the South African sub-cohort, and by 9\% in sensitivity (81\% sensitivity at specificity 62\%) when tested on the Ugandan sub-cohort.
The minimum specificity requirements are achieved for both the sub-cohorts.
Therefore, among the three considered architectures, LR shows merit as a potential screening tool, producing values which are consistently closer to the WHO minimum requirements.


\section{Conclusion}

We have evaluated the effect of country domain shift on the performance of neural network tuberculosis (TB) classification based on clinical and demographic data.
Experiments were performed using two sub-cohorts of the CAGE-TB dataset, collected in Uganda and South Africa respectively, which were used to train three classifier architectures: logistic regression, multilayer perceptron, and convolutional neural network.

Overall, the results show that such classifiers are robust to the domain shift associated with a change in the country from which the testing population is drawn.
The simple linear classifier (LR) was shown to achieve AUROCs of 0.8 [0.75,0.86] and 0.84 [0.78,0.9] when testing on the held-out Ugandan and South African data respectively.
The same model afforded the most consistent performance when evaluated at an operating point determined on the development set to achieve 90\% sensitivity, achieving test set sensitivity of 0.81 and specificity of 0.62 for Uganda and a sensitivity of 0.86 and specificity of 0.66 for South Africa.
This narrowly misses the WHO minimum requirements by 4-9\% in sensitivity even though the network is being evaluated on a completely held-out cohort.

Our investigation represents the first report of an evaluation of such classifiers across different countries, and indicates that TB triage and screening based on the automatic classification of clinical and demographic data is potentially viable. 
Future work will include the verification of these first results using larger and more varied cohorts whose heterogeneity may further validate the techniques applied here.


\section{Ethics and Data Availability}

This study is approved by the Health Research Ethics Committee of Stellenbosch University (M21/10/022), City of Cape Town (9517), and Western Cape Department of Health (WC\_202203\_035).

\section{Acknowledgements}
We acknowledge funding from the EDCTP2 program supported by the European Union (RIA2020I-3305, CAGE-TB). This project is also supported by Global Health EDCTP3 and its members (101145817, 4-CAGE-TB).
We gratefully acknowledge financial support from Telkom (South Africa) for the research presented in this paper.
Computations were performed using the University of Stellenbosch's HPC1 (Rhasatsha).

\section*{ORCID}
\small
\newcommand{\orcidline}[2]{%
  \textit{#1} & \href{https://orcid.org/#2}{%
    \includegraphics[width=1em]{orcid.pdf} orcid.org/#2}\\}
\noindent\begin{tabular*}{\linewidth}{@{}l@{\extracolsep{\fill}}r@{}}
\orcidline{Joshua M. Jansen van V\"{u}ren}{0000-0002-3406-4788}
\orcidline{Devendra S. Parihar}{0000-0001-6404-6428}
\orcidline{Daphne Naidoo}{0000-0001-7711-9526}
\orcidline{Marisa Klopper}{0000-0002-9318-8289}
\orcidline{Frank Cobelens}{0000-0002-4367-1133}
\orcidline{Lutz Kolbe}{0000-0003-4852-0040}
\orcidline{Kimsey Zajac}{0009-0005-0385-185X}
\orcidline{Willy Ssengooba}{0000-0002-1643-110X}
\orcidline{Moses Joloba}{0000-0002-0334-9983}
\orcidline{Grant Theron}{0000-0002-9216-2415}
\orcidline{Thomas R. Niesler}{0000-0002-7341-1017}
\end{tabular*}

\newpage

\bibliography{references}

\end{document}